\documentclass[runningheads]{llncs}
\usepackage{xcolor}
\usepackage{amsmath}
\usepackage[T1]{fontenc}
\usepackage{CJKutf8}
\usepackage{booktabs}
\usepackage{bm}
\usepackage{multirow}
\usepackage{tabulary}
\usepackage{array}
\usepackage{adjustbox}
\usepackage{courier}
\usepackage{color}
\usepackage{graphicx}
\begin{document}
\title{Overview of the NLPCC 2024 Shared Task on Chinese Metaphor Generation}
%
%

\newcommand{\equalcontribution}{\textsuperscript{*}}
\author{
  Xingwei Qu\textsuperscript{1,4}\thanks{Equal contribution} \and
  Ge Zhang\textsuperscript{2,3,4}\equalcontribution \and
  Siwei Wu\textsuperscript{1,4} \and
  Yizhi Li\textsuperscript{1,4} \and
  Chenghua Lin\textsuperscript{1,4}
}

\authorrunning{Xingwei Qu et al.}
%
\institute{University of Manchester \and
Stardust.AI \and
Univeristy of Waterloo \and Multimodal Art Projection Research Community}

\maketitle              
\begin{abstract}
This paper presents the results of the shared
task on Chinese metaphor generation, hosted at the 13th CCF Conference on Natural Language Processing and Chinese Computing (NLPCC 2024). The goal of this shared task is to generate Chinese metaphors using machine learning techniques and effectively identifying basic components of metaphorical sentences.  It is divided into two subtasks: 1) Metaphor Generation, which involves creating a metaphor from a provided tuple consisting of TENOR, GROUND, and VEHICLE. The goal here is to synthesize a metaphor that connects the subject (i.e. TENOR) with the object (i.e. VEHICLE), guided by the concept of the GROUND. 2) Metaphor Components Identification, which extracts the most fitting TENORs, GROUNDs, and VEHICLEs from a metaphorical sentence. This component requires the identification of 
 the most fitting metaphor elements that correspond to the specified grounds. In addition to overall results, we report on the setup and insights from the metaphor generation shared task, which attracted a total of 4 participating teams across both subtasks.

\keywords{metaphor generation  \and metaphor components identification}
\end{abstract}

\section{Introduction}
Metaphors play a crucial role in enhancing communication and description in everyday life. Their complexity necessitates sophisticated processing capabilities, making metaphor generation a significant area of research in natural language processing (NLP). Studies suggest that generating metaphors can improve various NLP applications, including creative language generation \cite{li2022cm}, sentiment analysis \cite{li-etal-2022-secret}, and machine translation \cite{mao-etal-2018-word}. The advent of large language models (LLMs) has marked notable advances in both metaphor generation and detection, with research such as that by Li et al. \cite{li2024findingchallengingmetaphorsconfuse} focusing on the challenges LLMs face in understanding metaphors.

The development of Chinese NLP corpora \cite{bai2023qwentechnicalreport} \cite{zhang2023corgipmchinesecorpusgender} and benchmarks \cite{wang2024mmtecorpusmetricsevaluating} 
\cite{zhang2023corgipmchinesecorpusgender} \cite{wang2023interactivenaturallanguageprocessing}\cite{ai2024yiopenfoundationmodels} has gained popularity. Shao et al. \cite{shao-etal-2024-cmdag-chinese} pioneered the creation of a Chinese metaphor dataset, highlighting significant challenges in Chinese metaphor generation.

In metaphor identification, word embeddings have become popular for capturing broad semantic relationships \cite{tsvetkov-etal-2014-metaphor}\cite{shutova-2015-design}\cite{rei-etal-2017-grasping}. Mao et al. \cite{mao-etal-2018-word} introduced an unsupervised learning approach targeting metaphor-expressing words, improving detection precision. Recently, Transformers \cite{attention} have been widely applied to metaphor detection tasks \cite{wang-etal-2023-metaphor}\cite{li-etal-2023-framebert}. Li et al. \cite{li-etal-2023-metaphor} emphasize the importance of understanding explicit meanings for effective metaphor detection.

The first subtask for this shared task for focuses on generating metaphorical sentences. This task aims to stimulate increased research interest in metaphor generation by spurring the development of innovative models and the construction of high-quality datasets. Specifically, for Chinese metaphor generation, the objective is to generate metaphors using only the TENOR and VEHICLE and their relationships, guided by the GROUND to intuitively craft metaphorical sentences. The second subtask involves using LLMs to identify each component of the given metaphorical sentences, thereby enhancing their information summarization capabilities. Ultimately, this initiative seeks to make technical texts more accessible to nonspecialist audiences and to advance the development of more user-friendly and effective metaphor generation and understanding models tailored to the Chinese linguistic domain, serving users with varying levels of expertise.

In this paper, we present the results of the Metaphor generation shared task, hosted by NLPCC 2024. We cover task description, datasets and shared task submission, before providing a description of participating systems, overall results anyalysis, and notable insights.

\section{Task Description}
The shared task is composed of two separate subtasks, focusing on 1) Creating a metaphorical sentence using provided components: TENOR, GROUND, and VEHICLE. 2) Identifying and extracting the fundamental elements of a metaphor (TENOR, GROUND, and VEHICLE) from a given metaphorical sentence.

\subsection{Subtask 1: Metaphor Generation}

The objective of this subtask is to train LLMs or develop rule-based methods, or utilize a combination of both to generate most fitting metaphor sentences with given words. Participants are then required to select the most accurate interpretation from four provided choices (A, B, C, D). For this purpose, we provide an example dataset of Chinese metaphors from \cite{shao-etal-2024-cmdag-chinese}, along with 500 training examples and an equal number of test examples which is self-curated in a multiple-choice format. This format not only allows the use of rule-based methods but also facilitates the application of large language models (LLMs) to generate predictions. This subtask is divided into two tracks:

\noindent\textbf{Track 1:} Submissions must employ large language models (LLMs) or System-2 LLMs. These models can incorporate additional modules such as extractive critics or classifiers to aid in metaphor generation.

\noindent\textbf{Track 2:} This track emphasizes the use of rule-based or classification-based machine learning methods without the involvement of LLMs. The focus is on harnessing structured algorithms and conventional ML techniques to interpret metaphors.

\subsection{Subtask 2: Metaphor components identification}

The goal of this subtask is to identify and extract the primary elements of metaphors—TENORs, GROUNDs, and VEHICLEs—from metaphorical sentences. Participants are required to develop methodologies that can accurately detect and classify these metaphor components. The TENOR represents the subject to which attributes are ascribed; the VEHICLE conveys the attribute in the metaphor; and the GROUND describes the basis of the relationship between the TENOR and the VEHICLE. This subtask is divided into two tracks:

\noindent\textbf{Track 1:} Submissions must employ Large Language Models (LLMs) or System-2 LLMs. These models can incorporate additional modules such as extractive critics or classifiers to aid in metaphor generation.

\noindent\textbf{Track 2:}  Participants must choose to utilize either rule-based systems or machine learning classifiers for this track. The rule-based approach involves the formulation of explicit rules that determine the relationships and components within the metaphor based on linguistic patterns. The machine learning approach, on the other hand, involves training models on annotated datasets to recognize and predict the metaphor elements.

\section{Dataset}

The dataset is divided into two parts. The first is a metaphor training dataset derived from the work of Shao et al. \cite{shao-etal-2024-cmdag-chinese}. This dataset, the largest of its kind, introduces Grounds into Chinese metaphor generation, highlighting their significance. For the second part which is used for this competition, Chinese metaphor sentences are sampled from two sources: Search engines are employed to find prominent examples using keywords such as "excellent metaphor sentences" and "classic metaphor sentences."
GPT-4 \cite{openai2024gpt4technicalreport} is used to identify renowned poetry and classic literature. Prompts are then crafted to extract metaphor sentences from these specific texts. 

We sample 1500 metaphorical sentences to construct this task, the source distribtion shown in Figure \ref{fig:two_images}. In subtask 2, we deconstruct these 1500 sentences into components, then use these sentences as the ground truth for our task 1. Additionally, we use these components to generate new sentences with GPT-4 \cite{openai2024gpt4technicalreport} and manual efforts, serving as distractors for subtask 1. For subtask, we have the following requirements:

\subsection{Metaphor Generation Curation Requirements for Subtask1}

We streamline the data annotation for Subtask1 with these refined criteria:

\noindent\textbf{Correct Answer:}
\begin{enumerate}
    \item Ensures accurate relationships among TENOR, VEHICLE, and GROUND.
    \item Metaphors are elegantly or wisely expressed.
\end{enumerate}

\noindent\textbf{Distractors:}
\begin{enumerate}
    \item Misalignment or misidentification of TENOR and VEHICLE.
    \item Incorrectly portrayed GROUND.
    \item Language lacks inspiration and charm.
\end{enumerate}

\noindent\textbf{Variation in Answer Length:}
Answers vary in length to prevent predictability, enhancing the challenge in identifying correct responses.

\noindent\textbf{Construction of Erroneous Options:}
Distractors should be grammatically correct yet lack excellence or elegance, with variable lengths to avoid simplicity.

\noindent\textbf{Strategic Use of Tenor and Vehicle:}
Swapping TENOR and VEHICLE positions or creating new elements from existing words enhances the depth and complexity of erroneous options.

\subsection{Metaphor Components Identification Curation Requirements for Subtask2}

We design the data annotation pipeline for Subtask2 based on the following condensed criteria:

\noindent\textbf{Ensuring Accuracy of TENOR, VEHICLE, and GROUND}
Accurately identify and describe the TENOR, VEHICLE, and similarities (GROUND) to ensure a precise representation of the metaphor.

\noindent\textbf{Handling Multiple VEHICLES}
Include all VEHICLES in cases with multiple metaphorical elements to capture the full complexity of the metaphor.

\noindent\textbf{Designing Diverse erroneous options}
Craft diverse erroneous options by:
\begin{enumerate}
    \item \textbf{Swapping TENOR and VEHICLE:} For instance, swapping ``middle-aged person'' with ``autumn rain.''
    \item \textbf{Using Alternative Nouns:} E.g., ``bird'' instead of ``birdcage'' in metaphors about marriage.
    \item \textbf{Varying Error Count:} Introduce distractors with varying error counts to obscure patterns.
    \item \textbf{Employing Synonyms:} Use synonyms to enhance the variability of distractors.
\end{enumerate}

\noindent\textbf{Correcting Original Sentences}:
Ensure the textual integrity by:
\begin{enumerate}
    \item \textbf{Eliminating Errors:} Correct all typographical and punctuation mistakes.
    \item \textbf{Simplifying Text:} Reduce complexity by focusing on a single TENOR.
    \item \textbf{Excluding Non-metaphorical Content:} Remove sentences that do not meet metaphorical criteria.
\end{enumerate}

\subsection{Dataset Pile}
\begin{figure}[h]
    \centering     \includegraphics[width=0.8\textwidth]{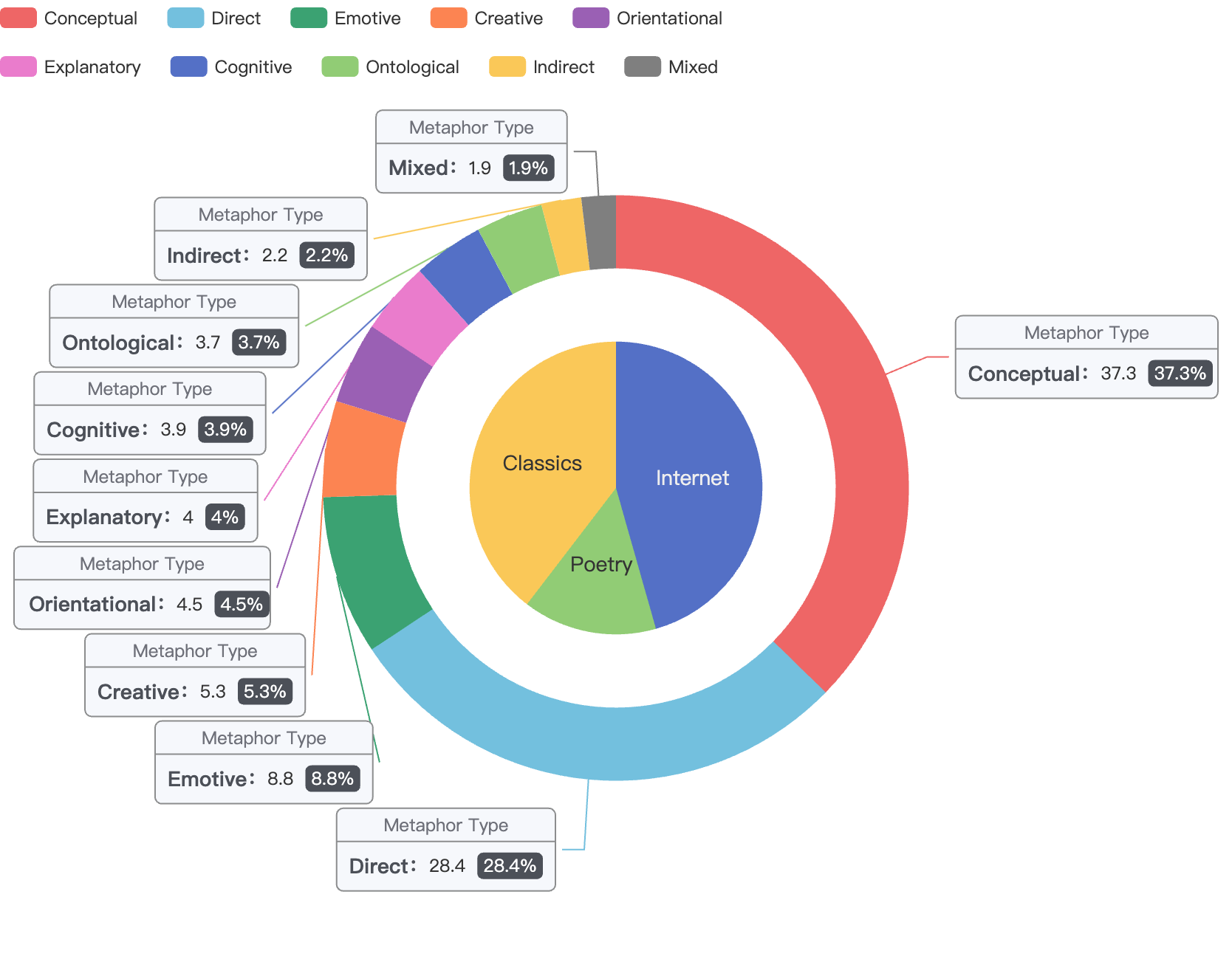}
    \caption{The outer ring represents the distribution of metaphor sentence types, while the inner ring depicts the distribution of sources}
    \label{fig:two_images}
\end{figure}
We divide each dataset containing 1,500 sentences into three equal parts, each comprising 500 questions. For validation, we provide 500 multi-choice examples along with the correct answers. The final evaluation is structured into two sections: the first section(Test A) contains 500 questions without provided answers; the second section(Test B) consists of 500 questions that are not visible to participants. Their submission code will be used to evaluate these hidden questions.

Figure \ref{fig:two_images} shows the categories about our metaphorical sentences type. Our categories are classified by GPT-4 \cite{openai2024gpt4technicalreport} and the prompt are shown in Appendix \ref{Prompt of labeling}. The detailed explanations of each metaphor category are shown in Figure \ref{tab:metaphor_categories}. 

\begin{table}[h]
\centering
\begin{tabular}{>{\raggedright\arraybackslash}p{2cm} >{\raggedright\arraybackslash}p{9.5cm}}
\toprule
\textbf{Metaphor Type} & \textbf{Description} \\ 
\midrule
Conceptual  & Metaphors where one concept is understood in terms of another. \\
Orientational  & Metaphors involving spatial orientation, like up/down, in/out. \\
Ontological & Metaphors where abstract concepts are represented as physical entities or substances. \\
Direct  & Metaphors that are explicitly stated in the language. \\
Indirect  & Metaphors that are implicit and require contextual interpretation. \\
Mixed  & Use of multiple metaphorical concepts within the same expression. \\
Explanatory  & Metaphors used to clarify or explain complex ideas. \\
Emotive & Metaphors used to express emotions or attitudes. \\
Cognitive  & Metaphors that facilitate thinking and processing of information. \\
Creative  & Highly original or novel metaphors that break conventional thinking patterns. \\
\bottomrule
\end{tabular}
\caption{Categories of Metaphors and Their Descriptions}
\label{tab:metaphor_categories}
\end{table}

\section{Shared Task Submissions}

\subsection{Team solution}

Our task attracted a total of 4 participating teams,
between them making a total of 32 submissions. A
brief explanation of the modelling approach taken
by each team is given below:

\noindent\textbf{KangGreen}: This team utilized the Yi-1.5-9b-chat model\cite{ai2024yiopenfoundationmodels}, renowned for its capabilities in Chinese text generation, for both metaphor-related subtasks. For the Metaphor Generation task, they fully fine-tuned the model using the high quality dataset set as Supervised Fine-Tuning (SFT) \cite{Ouyang2022TrainingLM} data, ensuring it adhered to specific instructions and queries. Similarly, in the Metaphor Components Identification task, they employed the same base model and fine-tuning approach, but modified the training prompts to suit the unique requirements of this subtask.

\noindent\textbf{ShaunTheSheep}: This team explores how high-quality data enhances large language models for metaphor generation. The team compiled several Chinese metaphor production datasets, streamlined by removing redundancies and using large language models to integrate elements uniformly. They also manually refined the datasets to improve quality and relevance. 

\noindent\textbf{YNU-HPCC}:
This team use a two-stage approach. Initially, the DeBERTa \cite{he2021debertav3} model generates a list of answer candidates, presented in brackets alongside their confidence scores. Subsequently, these candidates are combined with demonstrations to create a heuristic-enhanced prompt, refining the selection process based on the identified patterns and insights.

\noindent\textbf{ZZU-NLP}: This team developed a framework combining context-aware language learning with data augmentation to improve metaphor component identification. They processed initial data using ChatGPT\cite{openai2024gpt4technicalreport}, followed by supervised fine-tuning to refine the dataset. Additionally, they incorporated extra metaphor datasets for pre-training and utilized a Graph Attention Network (GAT) \cite{veličković2018graphattentionnetworks} encoder to efficiently retrieve contextual examples. Their methodology demonstrated significant improvements in identifying metaphor components.

\subsection{Benchmarking Models Performance}

Table \ref{tab:model_performance}  presents performance metrics of several large language models (LLMs) on tasks critical for evaluating Chinese language processing capabilities, particularly in metaphor generation. These models include Chinese Tiny LLM (CT-LLM) \cite{du2024chinese} (2B), MAP-NEO \cite{zhang2024mapneo} (7B), Yi-1.5-34B \cite{ai2024yiopenfoundationmodels}, Qwen2-72B \cite{bai2023qwentechnicalreport}\cite{yang2024qwen2technicalreport}, and GPT-4-Turbo \cite{openai2024gpt4technicalreport} (200B+), which vary significantly in computational capacity.

A notable observation is that although model size correlates with performance improvements, the gains diminish substantially beyond 34 billion parameters. This plateau suggests a potential upper bound in model capability for metaphor understanding tasks. Therefore, enhancing data quality might be essential to further improvements. These findings underscore the significance of our competition, which highlights persisting challenges in metaphor comprehension within advanced LLMs.

\section{Leadboard}

Table \ref{tab:leaderboard} displays the final leaderboard rankings for this competition, highlighting the winning teams in each subtask and track. In Subtask 1 Track 1, the top-performing team was KangGreen, utilizing the Yi-1.5-9b-chat \cite{ai2024yiopenfoundationmodels} model, fine-tuned with high-quality datasets following the Supervised Fine-Tuning (SFT) approach for both the Metaphor Generation and Metaphor Components Identification tasks. In Subtask 2 Track 1, the winning team was YNU-HPCC, which used a two-stage strategy with the DeBERTa \cite{he2021debertav3} model to generate answer candidates with confidence scores, refined through heuristic-enhanced prompts based on identified patterns and insights.

In summary, the results indicate that metaphor generation requires advanced fine-tuning techniques and robust dataset construction, while metaphor components identification demands careful elimination of erroneous options and a comprehensive understanding of the overall metaphor.

\begin{table}[ht]
\centering
\begin{minipage}{0.48\textwidth}
    \centering
    \caption{Performance of Different Models on Various Tests}
    \label{tab:model_performance}
    \resizebox{\textwidth}{!}{
    \begin{tabular}{>{\raggedright\arraybackslash}p{2.4cm}cccc}
    \toprule
    \textbf{Model Name} & \multicolumn{2}{c}{\textbf{Subtask 1}} & \multicolumn{2}{c}{\textbf{Subtask 2}} \\
    \cmidrule(lr){2-3} \cmidrule(lr){4-5}
    & \textbf{Test A} & \textbf{Test B} & \textbf{Test A} & \textbf{Test B} \\
    \midrule
    CT-LLM     & 28.2 & 21.0 & 5.6 & 25.8 \\
    MAP-NEO   & 48.4 & 49.0 & 37.8 & 40.7 \\
    Yi-1.5-34B     & 83.6 & 75.4 & 87.2 & 88.8 \\
    Qwen2-72B       & 85.8 & \textbf{82.8} & \textbf{93.4} & \textbf{92.4} \\
    GPT-4-Turbo    & \textbf{89.8} & 81.8 & 90.2 & 87.8 \\
    \bottomrule
    \end{tabular}
    }
\end{minipage}%
\hfill
\begin{minipage}{0.48\textwidth}
    \centering
    \caption{Leaderboard Results for All Subtasks}
    \label{tab:leaderboard}
    \resizebox{\textwidth}{!}{
    \begin{tabular}{@{}clcc@{}}
    \toprule
    \textbf{Subtask \& Track} & \textbf{Team Name} & \textbf{Test A} & \textbf{Test B} \\
    \midrule
    Subtask 1 Track 1 & kangreen & 98.8 & 98.0 \\
    & ShaunTheSheep & 96.2 & 96.0 \\
    & YNU-HPCC & 96.6 & 95.2 \\ \hline
    Subtask 1 Track 2 & YNU-HPCC & 98.4 & 97.4 \\ \hline
    Subtask 2 Track 1 & YNU-HPCC & 96.6 & 93.6 \\
    & ZZU-NLP & 93.8 & 91.82 \\
    & ShaunTheSheep & 92.2 & 92.81 \\
    & kangreen & 92.8 & 91.8 \\ \hline
    Subtask 2 Track 2 & YNU-HPCC & 95 & 93.2 \\
    \bottomrule
    \end{tabular}
    }
\end{minipage}
\end{table}
\vspace{-1cm}
\section{Result Analysis}

\begin{figure}[h!]
  \centering
  \includegraphics[width=0.9\linewidth]{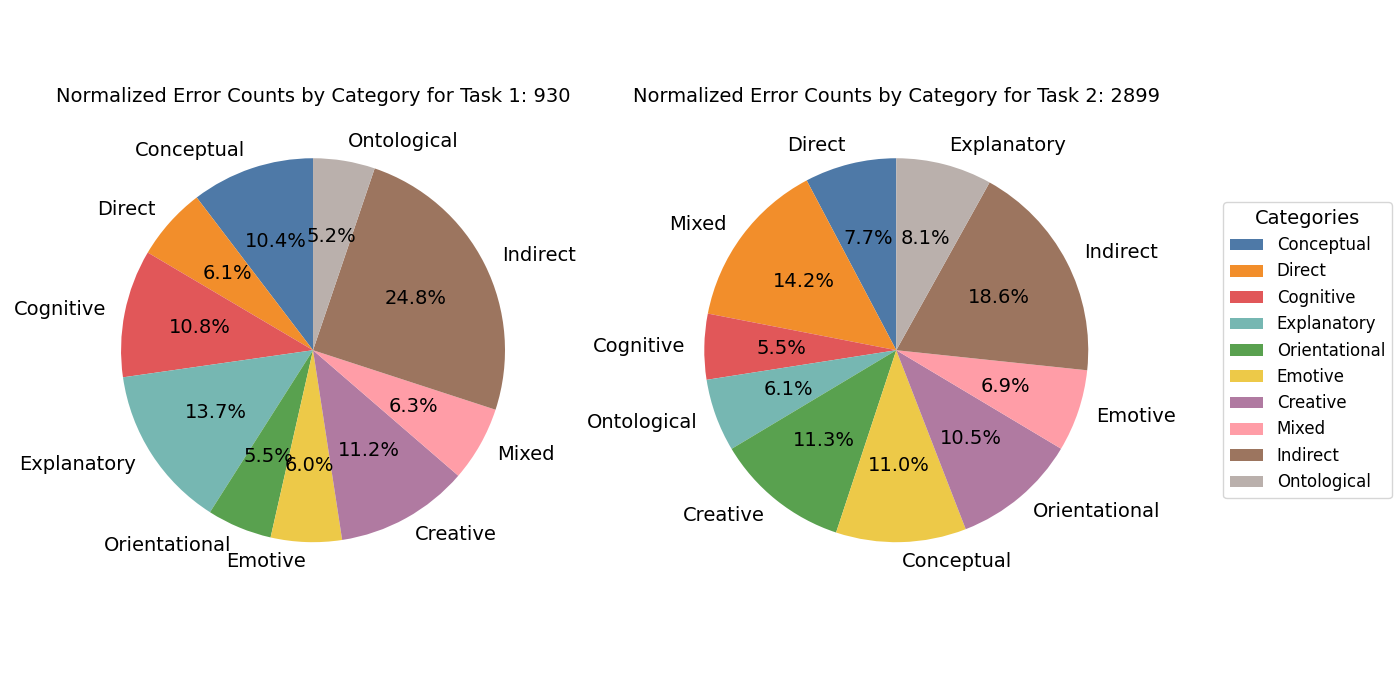}

  \caption{Normalized Error Counts by Category for Task 1 and Task 2. }
  \label{fig:metaphor_tasks}
\end{figure}

In Figure \ref{fig:metaphor_tasks}, the distribution of error types across two different tasks from all 4 teams submission highlights several insights into the challenge of understanding and generating metaphors. Notably, the category ``Indirect Metaphors'' shows a higher proportion of errors across both tasks, suggesting that teams struggle more with metaphors that require deeper contextual interpretation compared to those that are direct or explicit. Moreover, Task 2, which involves metaphor component identification, appears to be more challenging than Task 1, metaphor generation, as indicated by the higher error rates in several complex categories such as ``Indirect'' and ``Ontological'' metaphors. This reflects the intricate nature of deconstructing metaphors to their constituent components, a task that demands a more nuanced understanding of language. The normalization of error counts is calculated using the formula:
\[ \text{Normalized Error Count} = \frac{\text{Number of Errors in Category}}{\text{Proportion of Category in Dataset}} \]
This metric adjusts for the frequency of each metaphor type within the dataset, providing a fair comparison across different categories and highlighting which metaphor types are inherently more difficult to process correctly by the models.

This comparative analysis reveals that while metaphor generation is a complex task, the precise identification of metaphor components presents a greater challenge, underscoring the need for advanced linguistic models.

\vspace{-0.3cm}

\section{Conclusion}
This Chinese Metaphor Generation shared task was hosted
at the NLPCC2024 and consisted of two subtasks focusing on metaphor generation and metaphor components identification,
respectively. The task attracted a total of 4 teams,
between them making 32 individual submissions
across both subtasks. 
We provided comprehensive methods for the curation of Chinese Metaphor Generation. Our competitions demonstrate that even advanced models such as GPT-4-Turbo have limitations in fully understanding and generating metaphors. However, the majority of teams were able to fine-tune their Large Language Models (LLMs) with our curated dataset to achieve accuracies exceeding 90\%. Given these results, we believe that utilizing our data curation pipeline and resources will be a promising direction for future developments in Chinese metaphor generation.

\bibliography{custom}
\bibliographystyle{splncs04}

\appendix
\section{Prompt of Metaphor Categorization} \label{Prompt of labeling}
The prompt for our labeling prompts is :

You are a highly intelligent classification assistant who categorizes items accurately. I have some metaphor sentences, and I need you to provide the following information:

\noindent\textbf{Type}: The category of the metaphor, such as Conceptual, Orientational, Direct, etc. A metaphor sentence may fall into multiple categories, but you MUST provide ONE and ONLY ONE category that BEST fits the website.
Here is a detailed explanation of each metaphor category.

\noindent\textbf{Conceptual Metaphors}: Metaphors where one concept is understood in terms of another.

Orientational Metaphors: Metaphors involving spatial orientation, like up/down, in/out.

Ontological Metaphors: Metaphors where abstract concepts are represented as physical entities or substances.

Direct Metaphors: Metaphors that are explicitly stated in the language.

Indirect Metaphors: Metaphors that are implicit and require contextual interpretation.

Mixed Metaphors: Use of multiple metaphorical concepts within the same expression.

Explanatory Metaphors: Metaphors used to clarify or explain complex ideas.

Emotive Metaphors: Metaphors used to express emotions or attitudes.

Cognitive Metaphors: Metaphors that facilitate thinking and processing of information.

Creative Metaphors: Highly original or novel metaphors that break conventional thinking patterns.

\section{Benchmark Prompt} \label{benchamrk}

his section elaborates on the procedures for metaphor identification tasks facilitated by an open-sourced model.

\subsection{Subtask 1}
For Task 1, participants are prompted to identify the most appropriate metaphorical sentence based on the provided context and options. The instruction is structured as follows:

\begin{CJK*}{UTF8}{gbsn}基于本体和喻体，及其之间的共性,从下面隐喻句子的选项中，选出最符合的句子,请直接回答对应选项，不需要其他额外的回复:
\{text\}
\{options\}\end{CJK*}

In this structure, \texttt{\{text\}} represents the context, while \texttt{\{options\}} lists the available sentences for selection.

\subsection{Subtask 2}
In Subtask 2, participants are tasked with identifying the source and target concepts within a metaphorical context, alongside their shared properties. The format of this subtask is as follows:

\begin{CJK*}{UTF8}{gbsn}给定一段含有关于本体和喻体的隐喻(metaphor)的中文句子: \{text\}
请你找出其中的本体和喻体,以及本体和喻体间的共性:
\{options\}\end{CJK*}

Here, \texttt{text} is the metaphor-containing sentence, and \texttt{\{options\}} provides multiple-choice candidates.

\subsection{Post-Processing}
During post-processing, the initial letters 'A', 'B', 'C', and 'D' are extracted from the choices to ascertain participants' selections.

%





\end{document}